\documentclass[10pt,journal,compsoc]{IEEEtran}

\usepackage{times}
\usepackage{epsfig}
\usepackage{graphicx}
\usepackage{amsmath}
\usepackage{amsthm}
\usepackage{amssymb}
\usepackage{dsfont}
 \usepackage[ruled,vlined]{algorithm2e}
\usepackage{color}
\usepackage{soul}
\usepackage{caption}

\allowdisplaybreaks
\graphicspath{{fig/}}

\ifCLASSOPTIONcompsoc
  \usepackage[nocompress]{cite}
\else
  \usepackage{cite}
\fi

\usepackage[pagebackref=false,breaklinks=true,letterpaper=true,colorlinks,bookmarks=false]{hyperref}
\newif\ifdraft
\draftfalse
\drafttrue 

\definecolor{orange}{rgb}{1,0.5,0}
\definecolor{violet}{RGB}{70,0,170}
\definecolor{magenta}{RGB}{170,0,170}
\definecolor{dgreen}{RGB}{0,150,0}

\usepackage[normalem]{ulem}

\ifdraft
 \newcommand{\PF}[1]{{\color{red}{\bf PF: #1}}}
 
 \newcommand{\ShP}[1]{{\color{blue}{\bf SP: #1}}}
 
  \newcommand{\MS}[1]{{\color{dgreen}{\bf MS: #1}}}
 
\else
 \newcommand{\PF}[1]{}
 
 \newcommand{\ShP}[1]{}
 
 \newcommand{\MS}[1]{}
  
\fi

\newcommand{\comment}[1]{}
\newcommand{\parag}[1]{\vspace{0.5mm}\noindent {\bf #1}}

\newcommand{\bx}{\mathbf{x}}
\newcommand{\bX}{\mathbf{X}}

\usepackage{multirow}

\title{A Closed-Form Solution to Local Non-Rigid Structure-from-Motion}
\author{Shaifali~Parashar, Yuxuan~Long,
			Mathieu~Salzmann,
			Pascal~Fua,~\IEEEmembership{Fellow,~IEEE}%
	         \IEEEcompsocitemizethanks{
		\IEEEcompsocthanksitem S. Parashar, Y. Long, M. Salzmann and P. Fua are with the Computer Vision Laboratory, \'{E}cole Polytechnique F\'{e}d\'{e}rale de Lausanne, Switzerland. E-mail: \{shaifali.parashar,yuxuan.long,mathieu.salzmann, pascal.fua\}@epfl.ch.}}

\begin{document}

\markboth{IEEE TRANSACTIONS ON PATTERN ANALYSIS AND MACHINE INTELLIGENCEs}%
		{Shell \MakeLowercase{\textit{et al.}}: Bare Demo of IEEEtran.cls for Computer Society Journals}

\IEEEtitleabstractindextext{

A recent trend in Non-Rigid Structure-from-Motion (NRSfM) is to express local, differential constraints between pairs of images, from which the surface normal at any point can be obtained by solving a system of polynomial equations.  Unfortunately, these systems are of high degree with up to five real solutions. Hence, a computationally expensive strategy is required to select a unique solution. Furthermore, they suffer from degeneracies that make the resulting estimates unreliable, without any mechanism to identify this situation. 

In this paper, we show that, under widely applicable assumptions, we can derive a new system of equations in terms of the surface normals, whose two solutions can be obtained in closed-form and can easily be disambiguated locally. Our formalism also allows us to assess how reliable the estimated local normals are and to discard them if they are not. Our experiments show that our reconstructions, obtained from two or more views, are significantly more accurate than those of state-of-the-art methods, while also being faster.

}

\maketitle
\IEEEdisplaynontitleabstractindextext
\IEEEpeerreviewmaketitle


\section{Introduction}

Reconstructing the 3D shape of deformable objects from monocular image sequences is known as Non-Rigid Structure-from-Motion (NRSfM) and has applications in domains ranging from entertainment~\cite{Parashar19b} to medicine~\cite{Lamarca19}. Early methods relied on low-rank representations of the surfaces~\cite{Bregler00,DelBue06,Akhter08,Torresani01,Dai14,Lee16,Gotardo11a,Kumar19,Kumar19a}, while more recent ones exploit local surface properties to derive constraints and can handle larger deformations~\cite{Varol09,Vicente12,Taylor10a,Chhatkuli14,Chhatkuli17,Ji17}. Unfortunately, these constraints have to be enforced jointly on the entire set of reconstructed points for a whole sequence. Hence, the computational cost increases non-linearly with the number of images and quickly becomes prohibitive. Furthermore, a globally optimal solution is obtained using an iterative refinement, which requires a reliable initialization that is not always available.  Finally, these {\it global} methods cannot handle missing data.

In earlier work~\cite{Parashar17,Parashar19a,Parashar20}, we have shown that {\it local} methods constitute a powerful alternative. Expressing isometry, conformality, or equiareality constraints in terms of differential properties makes  the number of local variables remain fixed. Unfortunately, the systems of equations that arise in these computations are bivariate of high degree. They can have up to five real solutions. In theory, a unique solution can be obtained from 3 images, but this requires  either a 
complicated sum-of-squares formulation~\cite{Parashar17,Parashar19a} or reduction methodologies that add phantom solutions~\cite{Parashar20,Parashar21}. Hence, in practice, it takes more than 3 images to produce reliable estimates. Furthermore, when the motion between the frames is too small, the system becomes ill-posed and the estimates unreliable, without any mechanism to flag such situations as problematic.

In this paper, we introduce a new local method. Instead of inferring the  depth derivatives, we estimate surface normals. More specifically, given a 2D warp  between two images, we consider  tangent planes at corresponding points. For each pair of points, we compute the homography relating the two planes  and decompose it to compute the normals by solving local differential constraints~\cite{Parashar17,Parashar19a}. This has two solutions, instead of five in our earlier approaches~\cite{Parashar21}. For each plane, we pick the right one by enforcing an easy-to-compute measure of local smoothness. Furthermore, our formalism lets us assess how well-conditioned the  problem was and, hence, how usable the resulting normals are. In other words, we can derive from an image pair a set of reliable normals and discard the others. 

We will demonstrate on both synthetic and real data that we outperform state-of-the-art local and global methods at a fraction of the computational cost. Our contribution is therefore an approach to NRSfM that relies on solving in closed form a set of equations relating surface normals at corresponding points. Being entirely local, the computation is both fast and reliable.  Although our solution is designed for isometric or conformal deformations, it yields good results for generic ones. 



\begin{table*}[htpb]
\centering
\includegraphics[width=\textwidth]{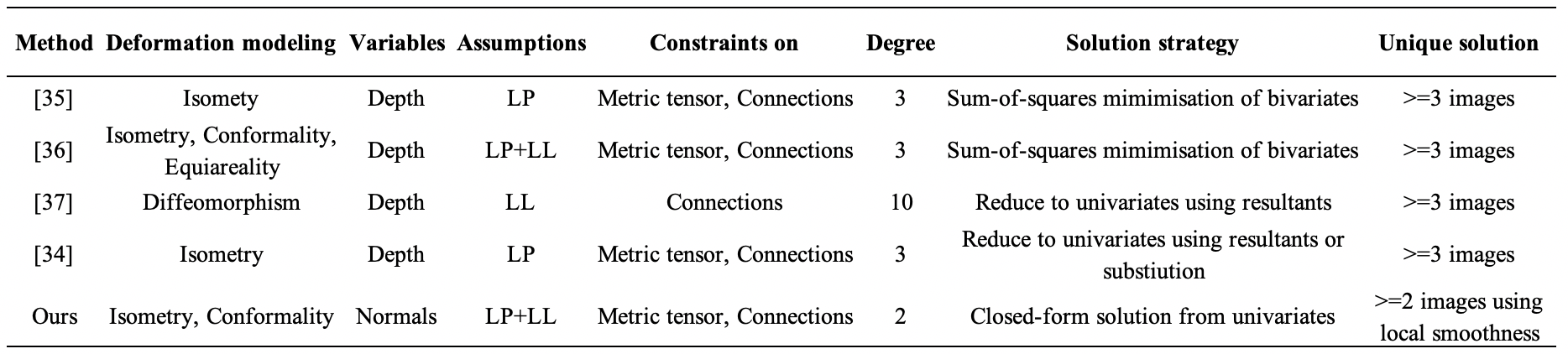}
\caption{Summary of the local methods we developed in earlier work.}
\label{table:summ}
\end{table*}
\section{Related Work}

NRSfM was introduced in~\cite{Bregler00} and the ill-posedness of the problem was handled by constraining the deformations to lie on a low-dimensional manifold.  Later variants introduced additional constraints for  efficient low-rank factorization ~\cite{DelBue06,DelBue08,Akhter08,Gotardo11a,Gotardo11b,Torresani08a} or performed additional optimization~\cite{Dai14,Paladini09,Garg13a,Lee16,Kumar18,Kumar19,Zhu14} to improve the statistical modeling.
Learning-based techniques have been used to tune the dimensionality of the deformation space~\cite{Kong19,Pumarola18,Golyanik18} using a large amount of annotated data for supervision.~\cite{Novotny19,Sidhu20} formulated learning-based techniques in an unsupervised setting to reconstruct  from sparse and dense data, respectively. However, this does not overcome the fundamental limitation of  approaches relying on low-rank assumptions: they cannot model complex deformations. Furthermore, they do not naturally handle missing data and occlusions, and complex formulations~\cite{Golyanik17} are required to overcome this.  As a result, these methods have been limited to objects that deform in a relatively predictable way, such as human faces. Recently, these limitations have been addressed by imposing constraints between corresponding points across images in one of the following ways.


\parag{Modeling Global Deformations.} Several methods seek to enforce physical  properties on the deformation, such as isometry that preserves local distances on the deforming surface. They approximate isometry by  inextensibility~\cite{Chhatkuli17,Ji17}, piece-wise inextensibility~\cite{Vicente12,Russell11,Russell14}, local or piece-wise rigidity~\cite{Taylor10a,Varol09,Chhatkuli14,Kumar19a}. A globally optimal solution is then found by jointly solving  over all corresponding points. This requires a computationally expensive optimization, which makes this approach impractical for handling large numbers of images. To handle non-isometric surfaces,  a mechanics-based approach is proposed in~\cite{Agudo14,Agudo15a,Agudo16}, introducing the forces required to compute the resulting shape. In any event, all these methods require an initialization, usually obtained using standard rigid-body reconstruction techniques. Furthermore, they are often  inaccurate.

\parag{Modeling Local Deformations.}  In earlier works, we have proposed methods that rely on formulating local deformation constraints in terms of algebraic expressions.  This makes it possible to reconstruct each surface point independently by solving algebraic equations, which reduces the computation cost. Being local, these methods inherently handle missing data and occlusions. In~\cite{Parashar17}, we treated surfaces as locally planar (LP) and formulated local isometric constraints using metric tensors  and connections representing the rate of change of metric tensors. In~\cite{Parashar19a}, we extended this deformation modeling to conformal and equiareal deformations by assuming the deformation to be locally linear (LL). For each pair of images, we obtained  two cubic equations in two variables  related to local depth derivatives with 9 possible solutions. In practice, up to 5 of them can be real. We found a unique solution by minimizing sum-of-squares of residuals over multiple images. In~\cite{Parashar21}, we proposed two fast solutions to the equations of isometric NRSfM~\cite{Parashar17}. 
Using substitution and resultants, we converted the original bivariate equations to univariate ones that can be solved efficiently. However, this comes at the cost of adding phantom solutions that cannot be identified. We picked the solution that yields the smallest residual of the isometry constraints on the entire image set. In~\cite{Parashar20}, we proposed an NRSfM solution for generic deformations. It uses only connections to formulate constraints to enforce surface smoothness. 

Table~\ref{table:summ} summarizes the characteristics of these local methods.
They tend to perform significantly better than their global counterparts but suffer from one key drawback: the local constraints are not always well-posed, leading to manyfold ambiguities or even degenerate solutions, without any mechanism for telling when this happens. 
This is the problem we address in this paper. 

\section{Formalism and Assumptions}
\label{sec:formalism}


\begin{figure}
\centering
\includegraphics[width=6cm]{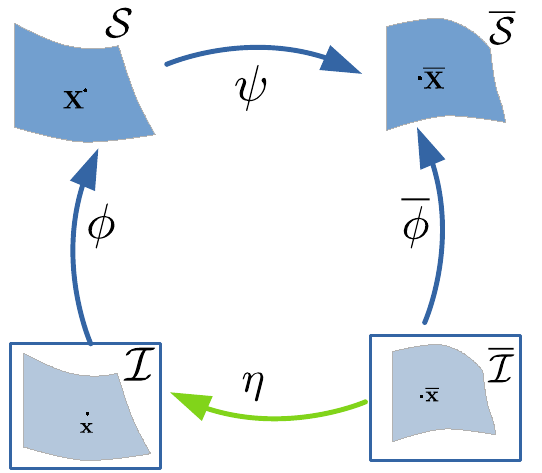}
\caption{A 2-view model for NRSfM. Assuming $\psi$ to be locally isometric/conformal, our goal is to find $\phi$, $\overline{\phi}$ given that $\eta$ is known. }
\label{fig:model}
\end{figure}

At the heart of our approach is the fact  that the normals at two different instants at a point on a deforming 3D surface can be computed given the point's projections in two images and a 2D warp between these images, under the sole assumptions of local surface planarity and deformation local linearity. In this section, we first introduce the NRSfM setup we will use in the rest of this paper, which is similar to the one of~\cite{Parashar19a}. We then explain what our assumptions mean and why they are widely applicable. Finally, we formulate the constraints we will use for reconstruction purposes. 

\subsection{Setup}

Fig.~\ref{fig:model} depicts our setup when using only two images, $\mathcal{I}$ and $\overline{\mathcal{I}}$, acquired by a calibrated camera. In each one, we denote the deforming surface as $\mathcal{S}$  and $\overline{\mathcal{S}}$, respectively, and model it  in terms of functions $\phi , \overline{\phi}: \mathbb{R}^2 \to \mathbb{R}^3$ that associate an image point to a surface point. Let us assume that we are given an image registration function $\eta: \mathbb{R}^2 \to \mathbb{R}^2$ that associates points in the first image to points in the second. 
This is often referred to as a warp. In practice, it can be computed using standard image matching techniques, such as optical flow~\cite{Sundaram10,Sun18a} or SIFT~\cite{Lowe04}. These functions can be composed to create a mapping $\psi: \mathbb{R}^3 \to \mathbb{R}^3$ between 3D surface points seen in the two images. We use a parametric representation of $\eta$ and $\phi$ with  B-splines~\cite{Bookstein89}, which allows us to accurately obtain first- and second-order derivatives of these functions. A finite-difference approach could also be used.

Given a point $\mathbf{x} = (u,v)$ on $ \mathcal{I}$ and its corresponding 3D point $\mathbf{X} = \phi(\mathbf{x})$  on $ \mathcal{S}$, we write $\phi(\mathbf{x})=\frac{1}{\beta(u,v)}\begin{pmatrix}
u & v & 1 \end{pmatrix}^\top$, where $\beta$  represents the inverse of the depth. 
The Jacobian of $\phi$ is given by
\begin{equation}
\label{eq:j_phi}
  \mathbf{J}_\phi = \frac{1}{\beta(u,v)}\begin{pmatrix}
1- uk_1 & -uk_2 \\ -vk_1 & 1-vk_2 \\ -k_1 & -k_2
\end{pmatrix},
\end{equation}
where  $k_1=\frac{\partial_u \beta}{\beta}$ and $k_2=\frac{\partial_v \beta}{\beta}$.  $\overline{u}$, $\overline{v}$, $\overline{\phi}$,  $\overline{k_1}$, and   $\overline{k_2}$ are defined similarly in $\overline{\mathcal{I}}$.
 
 \subsection{Local Planarity and Linearity}

In this work, we assume local planarity of the 3D surfaces and local linearity of the deformations as described in~\cite{Lee97,Kock10}. We now describe these two assumptions and argue that they are weak ones that are generally applicable.

\parag{Surface Local Planarity.}  
Let $\bx_0$ be an image point with surface normal $\mathbf{n}$ at $\phi(\bx_0)$. All points $\bx = (u,v)$ sufficiently close to $\bx_0$ can be accurately described as lying on the tangent plane. Hence, they satisfy $\mathbf{n}^\top\phi(\mathbf{x})+d=0$, where $d$ is a scalar, which we can rewrite as  $\beta = -\frac{\mathbf{n}^\top}{d}\begin{pmatrix} u & v & 1 \end{pmatrix}^\top$. Therefore the inverse depth $\beta$ that appears in Eq.~\ref{eq:j_phi} is a linear function of $\bx$ even though $\phi$ is not.  Nevertheless, all higher-order derivatives of $\phi$ can be expressed in terms of $\beta$ and its first-order derivatives. This is widely viewed as a weak assumption that applies to most smooth manifolds~\cite{Lee97}. For example, our planet is a sphere that can be treated as locally planar.

\parag{Deformation Local Linearity.} 
According to~\cite{Kock10}, every non-linear function can be approximated with an infinite number of linear functions. This assumption has been successfully used in shape-matching~\cite{Ovsjanikov12}. We assume the deformation $\psi$ that  relates locally two planes to be smooth enough to be well described locally by its first-order approximation, so that we can ignore its second derivatives.   In other words, we use a first-order approximation for the local deformations but a second-order one for the surface depth to allow for globally non-planar shapes.  This is a looser set of assumptions than what is normally used in NRSfM.  For example,~\cite{Lee16,Dai14} and other low-rank methods  assume the deformation space to be small; physics-based methods that use inextensibility~\cite{Chhatkuli17,Vicente12} or piecewise-rigidity~\cite{Varol09,Taylor10a} make a much stronger assumption.

Under the assumption of local planarity, we have $\mathbf{X}$ and $\overline{\mathbf{X}}$ lying on a planar surface. A generic transformation between these two surfaces, which defines the deformation $\psi$, can be expressed as $\overline{\mathbf{X}}= \mathbf{S}\mathbb{R}\mathbf{X} + \mathbf{T}$, where $\mathbb{R}$ and $\mathbf{T}$ are rotation and translation and $\mathbf{S}$ is a scaling matrix. If $\mathbf{S}$ happens to be a purely diagonal matrix with equal entries, $\psi$ is a planar homography, and the resulting deformation is purely isometric or conformal. Nevertheless, $\psi$ is linear. Therefore, local planarity of surfaces  implies local linearity of deformations. However, the reverse is not true.



\subsection{Differential Constraints across Images}
\label{sec:diffConstraints}

To express constraints between quantities computed in $\mathcal{I}$ and  $\overline{\mathcal{I}}$, we define {\it metric tensors} and {\it connections} as described in~\cite{Lee97}.

\parag{Metric Tensors.} 

The metric tensors $\mathbf{g}$ in $\mathcal{I}$ and $\overline{\mathbf{g}}$ in $\overline{\mathcal{I}}$ are first-order differential quantities that capture local distances and angles. They can be written as
 \begin{equation}
\mathbf{g}=\mathbf{J}_\phi^\top\mathbf{J}_\phi \text{ and } \overline{\mathbf{g}}=\mathbf{J}_{\overline{\phi}}^\top \mathbf{J}_{\overline{\phi}} \;,
\label{eq:nrsfm}
 \end{equation}
where $\mathbf{J}_\phi $ and $\mathbf{J}_{\overline{\phi}} $ are local surface jacobians computed according to Eq.~\ref{eq:j_phi}.
These tensors can be used to impose isometry, conformality, and equiareality constraints by forcing the scalars $k_1$ and $k_2$ of Eq.~\ref{eq:j_phi} to satisfy one of the three conditions below:
\begin{equation}
\label{eq:iso-nrsfm}
\begin{cases}
\overline{\mathbf{g}}\!=\!\mathbf{J}_\eta^\top \mathbf{g}\mathbf{J}_\eta,\!&\!\text{Isometry} 
\\
\overline{\mathbf{g}}\!=\!\lambda^2\mathbf{J}_\eta^\top \mathbf{g}\mathbf{J}_\eta, \! \lambda^2\!\in\!\mathbb{R}^{+}\! -\! \{1\},\!&\!\text{Conformality} 
\\
\sqrt{\det(\overline{\mathbf{g}})}\!=\!\sqrt{\det(\mathbf{J}_\eta^\top \mathbf{g}\mathbf{J}_\eta)},\!&\!\text{Equiareality} 
\end{cases}
\end{equation}
where $\mathbf{J}_\eta$ is the Jacobian of the warp $\eta$.

\parag{Linear Relation between Surface Derivatives.} 
Given $\mathbf{J}_\phi $, a local reference frame on the surfaces can be expressed with the column vectors as tangents and their cross product as normal. Connections are second-order differential quantities that express the rate of change of this local frame.
{Using connections under the assumption of local linearity as stated above, it can be shown~\cite{Parashar19a} that 
\begin{equation}
\label{eq:CS_rel}
\begin{pmatrix}
\overline{k}_1 \\ \overline{k}_2
\end{pmatrix} = \mathbf{J}_\eta^\top\begin{pmatrix}
k_1 \\ k_2
\end{pmatrix} -\begin{pmatrix}
0 & 1 \\ 1 & 0
\end{pmatrix}\mathbf{J}_\eta^{-1}\dfrac{\partial^2 \eta}{ \partial \overline{u}\overline{v}} ,
\end{equation}
where $\dfrac{\partial^2 \eta}{ \partial \overline{u}\overline{v}} $ are the second-order derivatives of the warp.
Solutions to isometric, conformal and equiareal NRSfM can be obtained by solving the metric tensor preservation equations in Eq.~\ref{eq:iso-nrsfm} under the constraints  of Eq.~\ref{eq:CS_rel}.


\section{Computing Normals from Two Images}

In earlier approaches~\cite{Parashar19a}, the  NRSfM problem was addressed by solving the system of Eq.~\ref{eq:iso-nrsfm} under the isometry, conformality, and equiareality constraints of Eq.~\ref{eq:iso-nrsfm} with respect to the variables $k_1$ and $k_2$ of Eq.~\ref{eq:j_phi}.  Here, we solve this system of equations directly in terms of the surface normals. We will show that, not only can this be done in closed form, but it also allows us to identify degenerate situations that result in unreliable estimates.

\parag{Differentiating the Warp.}

Let us consider a point $\overline{\mathbf{x}} = (\overline{u},\overline{v})^T$ in $\overline{\mathcal{I}}$ and its corresponding point $(u,v)^T=\eta(\overline{u},\overline{v})$  in  $\mathcal{I}$, with corresponding points on surfaces $\mathbf{X}$ and $\overline{\mathbf{X}}$. Assuming the surfaces to be locally planar means that there is a $3 \times 3$ homography matrix $\mathbf{H} = [h_{ij}]_{1 \leq i,j \leq 3}$ such that $\mathbf{X} = \lambda \mathbf{H}\overline{\mathbf{X}}$. Since we assume a perspective projection for the camera, we write
\begin{equation}
\label{eq:homography}
\mathbf{x} = \frac{1}{\overline{s}}\mathbf{H}\overline{\mathbf{x}} \implies
\!\begin{pmatrix}
u \\ v \\ 1
\end{pmatrix}\!=\!\frac{1}{\overline{s}}\!\begin{pmatrix}
h_{11} \!& \!h_{12} \!&\! h_{13} \\
h_{21} \!&\! h_{22} \!&\! h_{23} \\
h_{31} \!&\! h_{32} \!&\! h_{33} 
\end{pmatrix}\!\begin{pmatrix}
\overline{u} \\ \overline{v} \\ 1
\end{pmatrix}, \; 
\end{equation} 
where $\overline{s}=h_{31}\overline{u}\!+\!h_{32}\overline{v}\!+\!h_{33}$. The first- and second-order derivatives of $\eta$ can be computed as 
\begin{align}
\label{eq:eta_der}
\mathbf{J}_\eta = 
\begin{pmatrix}
\dfrac{\partial \eta}{\partial \overline{u}} & \dfrac{\partial \eta}{\partial \overline{v}}
\end{pmatrix}&=\frac{1}{\overline{s}}
\begin{pmatrix}
h_{11}\!-\!h_{31}u \!&\! h_{12}\!-\!h_{32}u \\
h_{21}\!-\!h_{31}v \!&\! h_{22}\!-\!h_{32}v
\end{pmatrix}  \; , \nonumber \\[1mm]
\begin{pmatrix}
\dfrac{\partial^2 \eta}{ \partial \overline{u}^2}  \! &  \! \dfrac{\partial^2 \eta}{ \partial \overline{u}\overline{v}}  \! & \! \dfrac{\partial^2 \eta}{ \partial \overline{v}^2}
 \end{pmatrix} 
&= -\frac{1}{\overline{s}} \mathbf{J}_\eta\begin{pmatrix}
2h_{31}  \! &  \! h_{32}  \!& \! 0 \\ 0  \!& \! h_{31} &2h_{32}
\end{pmatrix}.
\end{align}

\parag{Image Embedding and Local Normal.}

The unit normal $\mathbf{n}$ at $\mathbf{x}$ is the cross product of the columns of  the matrix $\mathbf{J}_\phi$ from~Eq.~\ref{eq:j_phi}. This lets us write
\begin{align}
\label{eq:normal}
\mathbf{n} & =\! \frac{1}{\beta^2\sqrt{\det \mathbf{g}}}\!\begin{pmatrix}
k_1 \\k_2 \\ 1\!-\!uk_1\!-\!vk_2
\end{pmatrix} \\
&=\!\frac{1}{\beta^2\sqrt{\det \mathbf{g}}}\!\begin{pmatrix}
\mathbf{I}_{2\times 2} & 0 \\ -\mathbf{x}^\top &1
\end{pmatrix}\!\begin{pmatrix}
k_1 \\ k_2 \\ 1
\end{pmatrix}\!. \nonumber \\
\Rightarrow 
\begin{pmatrix}
k_1 \\ k_2 \\ 1
\end{pmatrix} & =\!\beta^2\sqrt{\det \mathbf{g}}\begin{pmatrix}
\mathbf{I}_{2\times 2} & 0 \\ \mathbf{x}^\top &1
\end{pmatrix}\mathbf{n}. \label{eq:k1k2}
\end{align}
Given the normal $\mathbf{n}$ of Eq.~\ref{eq:normal}, we rewrite the matrix $\mathbf{J}_\phi$ of Eq.~\ref{eq:j_phi} as
\begin{align}
\label{eq:j_phi_in_n}
\nonumber
\mathbf{J}_\phi \!&\!=\! \frac{1}{\beta}\!\begin{pmatrix}
0 &  uk_1\!+\!vk_2\!-\!1 & k_2 \\
1\!-\!uk_1\!-\!vk_2 & 0 & \!-\!k_1 \\
\!-\!k_2 & k_1 & 0
\end{pmatrix}\!
\begin{pmatrix}
0 & 1 \\ \!-\!1 & 0 \\ v & \!-\!u
\end{pmatrix} \\
\!&\!=\! \beta\sqrt{\det \mathbf{g}}[\mathbf{n}]_{\times}\mathbf{E}.
\end{align}
We can now rewrite the differential constraints across images introduced in Section~\ref{sec:diffConstraints}  in terms of the normals.

\parag{Linear Relation between Surface Normals.} 

Given the $\eta$ derivatives from Eq.~\ref{eq:eta_der}, the linear relation of Eq.~\ref{eq:CS_rel} becomes
\begin{align}
\label{eq:CS_law}
\begin{pmatrix}
\overline{k}_1 \\ \overline{k}_2
\end{pmatrix}&=\!\mathbf{J}_\eta^\top\!\begin{pmatrix}
k_1 \\ k_2
\end{pmatrix}\!+\!\dfrac{1}{\overline{s}}\!\begin{pmatrix}
h_{31} \\h_{32}
\end{pmatrix} \; , \nonumber \\
\Rightarrow
\begin{pmatrix}
\overline{k}_1 \\ \overline{k}_2 \\ 1
\end{pmatrix}&=\!\begin{pmatrix} 
\mathbf{J}_\eta^\top & \mathbf{m} \\ 0 & 1 \end{pmatrix}\!\begin{pmatrix}
k_1 \\k_2 \\1
\end{pmatrix}\!. 
\end{align}
Using Eq.~\ref{eq:k1k2}, we rewrite the above expression  as
\begin{align}
\label{eq:CS_norm}
\nonumber
\overline{\mathbf{n}}\!&\!=\!\frac{\beta^2}{\overline{\beta}^2}\!\sqrt{\frac{\det \mathbf{g}}{\det \overline{\mathbf{g}}}}\mathbf{T}\mathbf{n}\\
\nonumber  \!&\!=\!\frac{\beta^2}{\overline{\beta}^2}\!\sqrt{\frac{\det \mathbf{g}}{\det \overline{\mathbf{g}}}}\!\begin{pmatrix}
\mathbf{I}_{2\times 2} & 0 \\ -\overline{\mathbf{x}}^\top &1
\end{pmatrix}\!\begin{pmatrix} 
\!\mathbf{J}_\eta^\top\!&\!\mathbf{m}\!\\ 0 & 1 \end{pmatrix}\!\begin{pmatrix}
\!\mathbf{I}_{2\times 2}\!&\!0\!\\ \mathbf{x}^\top &1
\end{pmatrix}\!\mathbf{n}
\\ \nonumber
\!&\!=\!\dfrac{ \beta^2}{\overline{s}\overline{\beta}^2}\!\sqrt{\frac{\det\!\mathbf{g}}{\det\!\overline{\mathbf{g}}}}\!\begin{pmatrix}
\!\mathbf{I}_{2\times 2}\!&\!0\!\\\!-\!\overline{\mathbf{x}}^\top\!&\!1\!
\end{pmatrix}\!\begin{pmatrix}
\!h_{11}\!&\!h_{21}\!&\!h_{31}\!\\
\!h_{12}\!&\!h_{22}\!&\!h_{32}\!\\
\!\overline{s}u\!&\!\overline{s}v\!&\!\overline{s}\!
\end{pmatrix}\!\mathbf{n}\! \\ &\!=\!\dfrac{ \beta^2}{\overline{s}\overline{\beta}^2}\sqrt{\frac{\det\!\mathbf{g}}{\det\!\overline{\mathbf{g}}}}\!\mathbf{H}^\top\!\mathbf{n} \; ,
\end{align}
which directly relates the two normals.

\parag{Metric Tensor.}

As shown in Fig.~\ref{fig:model}, we can write $\overline{\phi} = \psi \circ \phi \circ \eta$. Differentiating this expression and multiplying it by its transpose yields
\begin{equation}
\label{eq:mt_gen}
\overline{\mathbf{g}}=\mathbf{J}_{\overline{\phi}}^\top\mathbf{J}_{\overline{\phi}} = \mathbf{J}_\eta^\top\mathbf{J}_\phi^\top\mathbf{J}_\psi^\top\mathbf{J}_{\psi}\mathbf{J}_\phi\mathbf{J}_\eta.
\end{equation}

Using Eq~\ref{eq:j_phi_in_n}, we write $\mathbf{J}_\phi\mathbf{J}_\eta = \beta \sqrt{\det g}[\mathbf{n}]_\times \mathbf{E}\mathbf{J}_\eta$. Given the $\eta$ derivatives of Eq.~\ref{eq:eta_der}, we simplify $\mathbf{E}\mathbf{J}_\eta$ to $\frac{1}{\overline{s}}\begin{pmatrix}
\mathbf{h}_1 \times \hat{\mathbf{x}} & \mathbf{h}_2 \times \hat{\mathbf{x}}
\end{pmatrix}$, where $\mathbf{h}_1, \mathbf{h}_2$ are the first two columns of the homography matrix $\mathbf{H}$, and $\hat{\mathbf{x}} = \begin{pmatrix}
u & v & 1
\end{pmatrix}^\top$.
By writing $\mathbf{z}_1 = \mathbf{n} \times (\mathbf{h}_1 \times \hat{\mathbf{x}})$ and $\mathbf{z}_2 = \mathbf{n} \times (\mathbf{h}_2 \times \hat{\mathbf{x}})$, Eq.~\ref{eq:iso-nrsfm} reduces to
\begin{equation}
\label{eq:mt_norm_simp}
\begin{cases}
\overline{\mathbf{g}}\!=\!\frac{\lambda^2\beta^2\det (\mathbf{g})}{\overline{s}^2}\begin{pmatrix}
\mathbf{z}_1^\top \mathbf{z}_1 \!\! & \!\! \mathbf{z}_1^\top \mathbf{z}_2 \\ \mathbf{z}_1^\top \mathbf{z}_2 \!\! &\!\! \mathbf{z}_2^\top \mathbf{z}_2 
\end{pmatrix} \!,\!&
\\
\sqrt{\det(\overline{\mathbf{g}})}\!
=\sqrt{\det(\mathbf{J}_{\eta}^\top \mathbf{g}\mathbf{J}_{\eta})}.\!&
\end{cases}
\end{equation}
%
%

\parag{NRSfM from Isometric/Conformal Constraints.} 

 So far, we have expressed the metric preservation conditions in terms of the normals of the two surfaces under consideration. The only unknown left in the system is therefore $\mathbf{n}$. We now show that this unknown can in fact be computed in closed form.


Given the multiplicative 
nature of the cross product, the constraints on the normals of Eq.~\ref{eq:CS_norm} imply that  
\begin{equation}
[\overline{\mathbf{n}}]_\times=\dfrac{ \beta^2}{\overline{s}\overline{\beta}^2}\sqrt{\frac{\det\mathbf{g}}{\det\overline{\mathbf{g}}}}\det (\mathbf{H}^\top) \mathbf{H}^{-1}[\mathbf{n}]_\times \mathbf{H}^{-\top} \; .
\end{equation}
This lets us rewrite the matrix $\mathbf{J}_{\overline{\phi}}$ of Eq.~\ref{eq:j_phi_in_n} as
\begin{align}
\label{eq:jphi_n_overline}
\nonumber
\mathbf{J}_{\overline{\phi}}&\!=\!\frac{ \beta^2}{\overline{\beta}}\sqrt{\det \mathbf{g}} \mathbf{H}^{-1}\![\mathbf{n}]_\times\!\left(\!\frac{\det \mathbf{H}^\top}{\overline{s}}\!\mathbf{H}^{-\top}\overline{\mathbf{E}}\!\right)
\\
&\!=\!\frac{\ \beta^2\!\sqrt{\det \mathbf{g}}}{\overline{\beta}}\!\mathbf{H}^{\!-1}\![\mathbf{n}]_\times\!\begin{pmatrix}
\!\mathbf{h}_1\!\times\!\hat{\mathbf{x}}\!&\!\mathbf{h}_2\!\times\!\hat{\mathbf{x}}\!\end{pmatrix} \\ \nonumber
&\!=\!\frac{ \beta^2\!\sqrt{\det \mathbf{g}}}{\overline{\beta}}\!\mathbf{H}^{\!-1}\!\begin{pmatrix}
\!\mathbf{z}_1\!&\!\mathbf{z}_2\!
\end{pmatrix}\!.
\end{align}
Injecting this expression into  the isometric/conformal metric tensor preservation relation of Eq.~\ref{eq:mt_norm_simp} yields 
\begin{small}
\begin{align}
\label{eq:recon_eqn}
\begin{pmatrix}
\mathbf{z}_1^\top\mathbf{H}^{\!-\top}\!\mathbf{H}^{\!-\!1} \mathbf{z}_1\!&\!\mathbf{z}_1^\top\mathbf{H}^{\!-\!\top}\!\mathbf{H}^{\!-\!1} \mathbf{z}_2\! \\ \mathbf{z}_1^\top\mathbf{H}^{\!-\!\top}\!\mathbf{H}^{\!-\!1} \mathbf{z}_2\!&\!\mathbf{z}_2^\top\mathbf{H}^{\!-\!\top}\!\mathbf{H}^{\!-\!1} \mathbf{z}_2\end{pmatrix}&=\!\frac{\lambda ^2 \overline{\beta}^2}{\overline{s}^2\beta^2}\!\begin{pmatrix}
\mathbf{z}_1^\top \mathbf{z}_1  \!\!&  \!\!\mathbf{z}_1^\top \mathbf{z}_2 \\ \mathbf{z}_1^\top \mathbf{z}_2  \!\!& \!\! \mathbf{z}_2^\top \mathbf{z}_2 
\end{pmatrix}\!, \nonumber \\
\Rightarrow \mathbf{z}_i^\top \left(\overline{\mathbf{H}}^\top\overline{\mathbf{H}}-\frac{\lambda^2 \overline{\beta}^2}{\overline{s}^2\beta^2}\mathbf{I}_{3\times 3}\right)\mathbf{z}_j & = 0, \forall i,j \in \{1,2\},
\end{align}
\end{small}
where $\overline{\mathbf{H}}=\mathbf{H}^{-1}$. Assuming $\mathbf{H}$ to be normalized, that is, its second singular value to be 1, the relation between a 3D point observed in the two input images is given by $\phi(\mathbf{x}) = \mathbf{H} \phi(\overline{\mathbf{x}}) $. Using Eq.~\ref{eq:homography} yields $\overline{\beta} = \beta \overline{s}$. By writing $\mathbf{z}_i = [\mathbf{n}]_{\times}[\mathbf{h}_i]_\times \hat{\mathbf{x}}$, the above constraints further simplify to
\begin{equation}
\label{eq:recon_H}
[\mathbf{n}]_\times^\top (\overline{\mathbf{H}}^\top\overline{\mathbf{H}}-\lambda^2\mathbf{I}_{3\times 3})[\mathbf{n}]_\times = 0.
\end{equation}
Since $ \mathbf{H} \sim \alpha \mathbf{H},$  we divide the above expression by $\lambda^2$ and, with a slight abuse of notation, write $\frac{1}{\lambda}\overline{\mathbf{H}}$ as $\overline{\mathbf{H}}$.
This simplifies the above expressions to
\begin{equation}
\label{eq:final}
[\mathbf{n}]_\times^\top (\overline{\mathbf{H}}^\top\overline{\mathbf{H}}-\mathbf{I}_{3\times 3})[\mathbf{n}]_\times =[\mathbf{n}]_\times^\top \mathbf{S}[\mathbf{n}]_\times = 0.
\end{equation}

\parag{Degenerate Cases.} The system of Eq.~\ref{eq:final} holds as long as $\mathbf{S}$ is a non-null matrix, which means $\overline{\mathbf{H}}^\top \overline{\mathbf{H}} \neq \mathbf{I}_{3\times3}$. Therefore, $\overline{\mathbf{H}}$ should not be an orthogonal matrix, which makes pure rotations and reflections cause degeneracies.

\parag{Affine Stability.} Under affine imaging conditions, $h_{31}=h_{32}=0$, and $h_{33}=1$.
In this case, $\mathbf{z}_i$ and $\mathbf{S}$ remain  non-null, and thus the system in Eq.~\ref{eq:final} does \emph{not} become degenerate, and we can still compute the normal. 

\parag{Solution.} The solution to the system in Eq.~\ref{eq:final} can be obtained by homography decomposition~\cite{Malis07}. We give an overview of the solution here but recommend reading~\cite{Malis07} for more detail.

$\mathbf{S}=\{s_{ij}\}$ is a symmetric matrix expressed in terms of $\overline{\mathbf{H}}$. It can be numerically computed using $\eta$ and image observations $(\mathbf{x},\overline{\mathbf{x}})$.  Specifically, Eq.~\ref{eq:CS_norm} gives the closed-form definition 
$\mathbf{H}^\top = 
\begin{pmatrix}
\mathbf{I}_{2\times 2} & 0 \\ -\overline{\mathbf{x}}^\top &1
\end{pmatrix}\!\begin{pmatrix} 
\!\mathbf{J}_\eta^\top\!&\!\mathbf{m}\!\\ 0 & 1 \end{pmatrix}\!\begin{pmatrix}
\!\mathbf{I}_{2\times 2}\!&\!0\!\\ \mathbf{x}^\top &1
\end{pmatrix}$.  Let us write $\mathbf{n} = \begin{pmatrix}
n_1 & n_2 & n_3
\end{pmatrix}^\top$. Since $n_3 \neq 0$, we define $y_1 = \frac{n_1}{n_3}$ and $y_2 = \frac{n_2}{n_3}$ and expand the system in Eq.~\ref{eq:final} accordingly. This yields 6 constraints, out of which only 3 are unique. They are given by
\begin{align}
\label{eq:quad_recon}
\nonumber s_{33} y_2^2 - 2s_{23} y_2 + s_{22} &= 0\;, \\
\nonumber s_{33} y_1^2 - 2s_{13} y_1 + s_{11} &= 0\;, \\
s_{22} y_1^2 - 2s_{12} y_1 y_2 + s_{11} y_2^2 &= 0\;.
\end{align}
By solving the first two,
we obtain $y_1 = \dfrac{s_{13} \pm \sqrt{s_{13}^2-s_{33}s_{11}}}{s_{33}}$ and $y_2 =\dfrac{s_{23} \pm \sqrt{s_{23}^2-s_{33}s_{22}}}{s_{33}}$. We use the third expression to disambiguate the solutions. Ultimately, this gives us closed-form expressions
for the two potential solutions for the normal,
written as
\begin{align}
\label{eq:normal_sol}
\nonumber
&\mathbf{n}_a = \begin{pmatrix}
s_{13} + s\sqrt{s_{13}^2-s_{33}s_{11}}& s_{23} + \sqrt{s_{23}^2-s_{33}s_{22}}&s_{33}
\end{pmatrix}^\top\!, \\
\nonumber
&\mathbf{n}_b = \begin{pmatrix}
s_{13} - s\sqrt{s_{13}^2-s_{33}s_{11}}&s_{23} - \sqrt{s_{23}^2-s_{33}s_{22}}&s_{33}
\end{pmatrix}^\top\!, \\
&\text{where } s = sign(s_{23}s_{13}-s_{12}s_{33}).
\end{align}

\parag{Normal Validation.}  The normals thus obtained must be visible to the camera.  Given the analytical normal in Eq.~\ref{eq:normal}, $\mathbf{n}_a $ and $\mathbf{n}_b $ are visible if  $\frac{s_{33}}{1\!-\!uk_1\!-\!vk_2} >0$, i.e., they have a similar orientation towards the camera. We discard the normals that do not meet the visibility constraint.

\parag{Normal Selection.} Using Eq.~\ref{eq:k1k2}, the local depth derivatives $(k_1,k_2)$ at $\mathbf{X}$ are given by  $k_i=\frac{n_i}{u n_1 + vn_2 + n_3}$. 
 From the solution in Eq.~\ref{eq:normal_sol}, we thus obtain two possible solutions for the local depth derivatives
$(k_{1a},k_{2a})$ and $(k_{1b},k_{2b})$.
We pick the normal that minimizes the corresponding sum of squares of depth derivatives. That is, we compute the normal $\mathbf{n}$  as
\begin{equation}
\label{eq:nc}
\mathbf{n} = \begin{cases}
\mathbf{n}_a  & \text{ if } k_{1a}^2 + k_{2a}^2 \leq  k_{1b}^2 + k_{2b}^2 \!,\\
\mathbf{n}_b  & \text{ otherwise. } 
\end{cases}
\end{equation}
Following Eq.~\ref{eq:homography},  $\overline{\mathbf{n}}$ is then obtained as $\mathbf{H}^\top\mathbf{n}$.

\parag{Measure of Degeneracy.} In degenerate situations, the singular values $(\sigma
_1,\sigma_2,\sigma_3)$ of $\mathbf{H} $ are all one. 
We use the ratio $\frac{\sigma_1}{\sigma_3}$ to quantify the degeneracy. Thus, we only reconstruct from $\mathbf{S}$ if $\frac{\sigma_1}{\sigma_3} > \tau$, and we set $\tau = 1.05$. 

\parag{Surface Reconstruction.}  We consider a planar surface and bend it to match the normals obtained using the homography decomposition mentioned above, as opposed to~\cite{Parashar19a,Parashar20,Parashar21} which integrate the normals on each surface. The upside of surface bending is that it does not require to set a smoothness parameter, which needs to be tuned for the normal integration. Furthermore, surface bending is much faster than its normal integration counterpart in the presence of dense data. It is also less affected by the noise in the normals corresponding to high-perspective image regions.

\section{Normals from Multiple Images}

Methods such as those of~\cite{Parashar19a,Parashar20,Parashar21} pick a reference image and formulate reconstruction constraints between it and the other images, which are then solved by solving a least-squares problem over the entire set of images. We use the same strategy, except that we reconstruct from image pairs, one of them being the reference image. Therefore, for $N$ images, we obtain $N-1$ estimates for the reference image and $1$ estimate for each of the non-reference images.

More formally, let $\{\bx ^i_j\}, i\in [1,M], j\in[1,N]$, be a set of $N$ point correspondences between $M$ images. Our goal is to find the 3D point $\bX^i_j$ and the normal $\mathbf{n}^i_j$ corresponding to each $\bx ^i_j$. Using Eq.~\ref{eq:CS_norm}, we write the local homography for each point correspondence $\mathbf{H}^{ik}_j$ between image pairs $(i,k) \in [1,M], i \neq k$, using the warp $\eta$. Each local homography $\mathbf{H}^{ik}_j$ is normalized by dividing it by its second singular value. We compute $\mathbf{Hc}^{ik}_j$ given by the ratio of the first and third singular value, and the normals for each local homography $\mathbf{H}^{ik}_j$ using Eq.~\ref{eq:normal_sol}.  We then pick a unique solution using Eq.~\ref{eq:nc}. The solution on the reference and non-reference  image is given by $\mathbf{n}_j^{kk}$ and $\mathbf{n}_j^{ki}$, respectively. For non-degenerate cases, where $\frac{\sigma_1}{\sigma_3} \geq 1.05$, we compute the normal ${\bf n}^i_j$ by taking the median of the $\mathbf{n}_j^{ik}$s computed over $k$ reference images. We obtain a 3D surface by bending a planar surface to match the obtained normals on each surface.
We summarize our complete pipeline in Algorithm~\ref{alg:diff}.

\begin{algorithm}
\SetAlgoLined
 \KwData{$\bx^i_j, \mathbf{H}^{ik}_j$ and $\mathbf{Hc}^{ik}_j$}
 \KwResult{$\mathbf{n}^i_j$}
 $\frac{\sigma_1}{\sigma_3} = 1.05$\;
 \For{ each reference image $k = [1,M]$}{
 \For{ each point $j= [1,N]$}{  
 \For{  images $i= [1,M], i \neq k$}{
 \eIf{$\mathbf{Hc}^{ik}_j > \frac{\sigma_1}{\sigma_3}$}{
  Compute normals using~\eqref{eq:normal_sol}\;
  Pick a solution $\mathbf{n}_j^{kk}$ using~\eqref{eq:nc}\;
  Write $\mathbf{n}_j^{ik} = (\mathbf{Hc}^{ik}_j)^\top \mathbf{n}_j^{kk}$\;
 }
 { Set $\mathbf{n}_j^{ik}, \mathbf{n}_j^{kk}$ to zero\;
 }
 }
 }
 }
  \For{ each point $j= [1,N]$}{  
   \For{  images $i= [1,M]$}{
 Obtain $\mathbf{n}^i_j$ by as the median of the non-zero $\mathbf{n}^{ik}_j$s\;
 }
 }
 \caption{Our NRSfM Algorithm}
 \label{alg:diff}
\end{algorithm}



\begin{figure*}
    \centering
     \includegraphics[width=\textwidth]{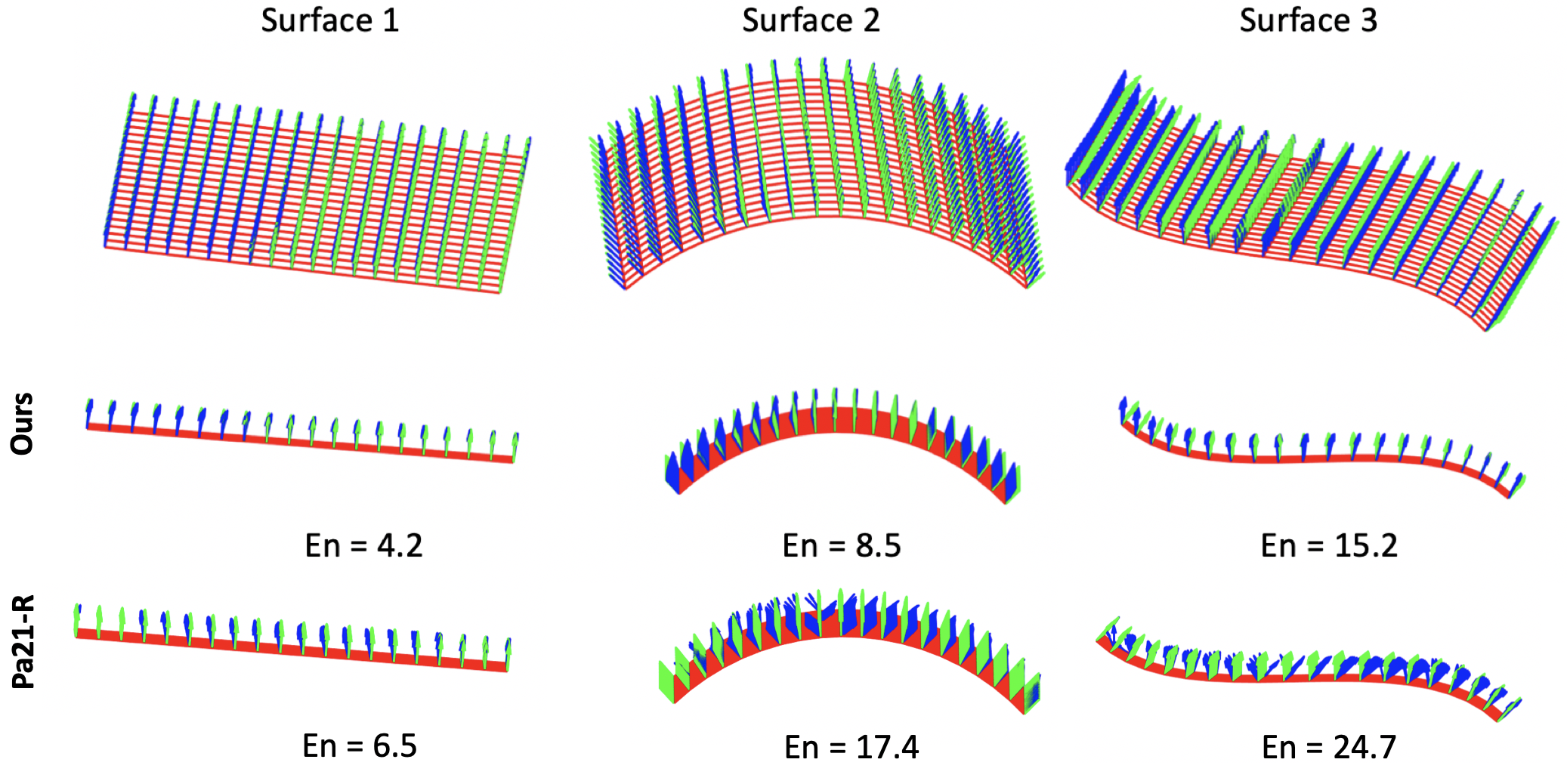}
     \caption{ {\bf Reconstructed normals}. A synthetic deforming surface reconstructed in three different frames. The predicted normals are shown in blue and the ground-truth ones in green.}
    \label{fig:syn_exp}
\end{figure*}

  \begin{figure*}
    \centering
     \includegraphics[width=\textwidth]{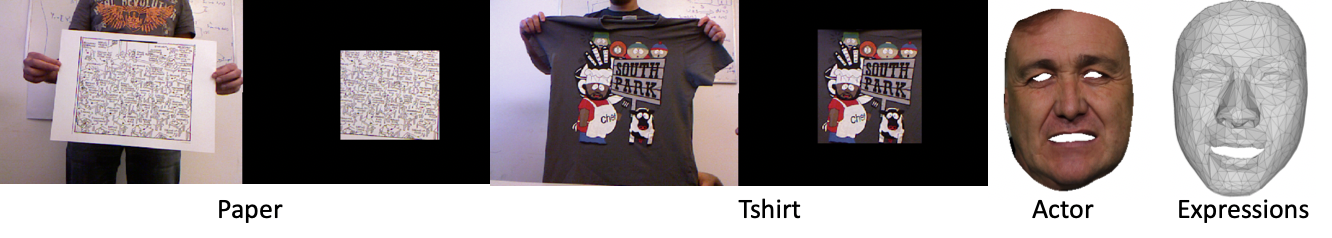}
     \caption{ {\bf Datasets used by~\cite{Sidhu20}.} Sample images and region of interest used for depth computation for \textbf{Paper}, \textbf{Tshirt} and \textbf{Actor} sequences. For \textbf{Expressions}, sample ground truth shape is shown. }
    \label{fig:dense_data}
\end{figure*}
 
\section{Experiments}

We compare our method against state-of-the-art ones on both synthetic and real datasets with available ground truth. 

\subsection{Datasets}

\parag{Synthetic Datasets.}
We created 3 smooth surfaces: a plane, a cylindrical surface and a  stretched surface with 400 tracked correspondences, as shown in Figure~\ref{fig:syn_exp}. 

\parag{Real Datasets from our Previous Work.}
These include the \textbf{Paper}~\cite{Salzmann07b}, \textbf{Rug}~\cite{Parashar17} and \textbf{Tshirt}~\cite{Chhatkuli14} datasets.  \textbf{Paper} comprises 191 images from a video of a deforming sheet of paper with 1500 point correspondences. \textbf{Rug} comprises 159 images from a video of a deforming rug with 3900 point correspondences. \textbf{Tshirt} has 10 wide-baseline images with 85 point correspondences.
The correspondences in the {\bf Paper} dataset were obtained using SIFT with a manual supervision of accuracy and are thus highly accurate. By contrast, those in the {\bf Rug} dataset were computed using the dense optical flow method of~\cite{Garg13a} and contain errors due to optical drift and regional mismatches due to the lack of texture. The correspondences in {\bf Tshirt} are computed manually. The ground truth for \textbf{Paper} and \textbf{Rug} is obtained using kinect, which is very noisy and contains large, inconsistent depth variations. For an apt comparison, we refined the ground truth to obtain smooth surfaces. The ground truth for {\bf Tshirt} is computed using rigid reconstruction of each image from multiple views.

\parag{NRSfM Challenge Dataset.} 
It consists of 5 image sequences depicted by Fig.~\ref{fig:nrsfm_ch_data}. They feature 5 kinds of non-rigid motions: articulated (piecewise-rigid) with 207 images and 69 point correspondences, balloon (conformal)  with 51 images and 211 point correspondences, paper bending (isometric) with 40 images and 153 point correspondences, rubber (elastic) with 40 images and 481 point correspondences, and paper being torn with 432 images and 405 point correspondences. The dataset features images from 6 different camera motions and provides image points captured assuming both a perspective and an orthographic projection. It provides only one ground-truth surface for each of the sequences.  The correspondences are sparse and not well-distributed across the images. 

\parag{Datasets used by~\cite{Sidhu20}.}
\cite{Sidhu20} released the \textbf{Paper}, \textbf{Tshirt}, \textbf{Actor} and \textbf{Expressions} datasets, which have been widely used by many physics-based and low-rank constraints based methods. The \textbf{Paper} images are the same as the one used by us.~\cite{Sidhu20} uses 60K dense correspondences computed using optical flow~\cite{Garg13a} and the raw depth data from the kinect is considered as the ground truth. The \textbf{Tshirt} data has 300 images with 70K dense correspondences computed using~\cite{Garg13a}, with the kinect raw depth data as ground truth. To deal with the inconsistent depth variations of the raw kinect data,~\cite{Sidhu20} refines the raw data and focuses on small portions of theses datasets where the inconsistent depth variations are minimal, as shown in Fig~\ref{fig:dense_data}.
\textbf{Actor} contains 100 images of a deforming human face with 36K dense correspondences,  and \textbf{Expressions} includes 384 3D shapes of a deforming human faces with 1000 point correspondences.  The ground truth for both these datasets is synthetic. Fig~\ref{fig:dense_data} shows some samples.

\parag{Blue Sheet Dataset.}
Additionally, we recorded a video sequence featuring a textureless blue sheet deforming isometrically using a Kinect. It comprises 60 images and 7K point correspondences that were tracked using  dense optical flow~\cite{Garg13a}. Optical flow on textureless surfaces is prone to large errors, and the flow we obtained confirms this. 

  \begin{figure*}
    \centering
     \includegraphics[width=\textwidth]{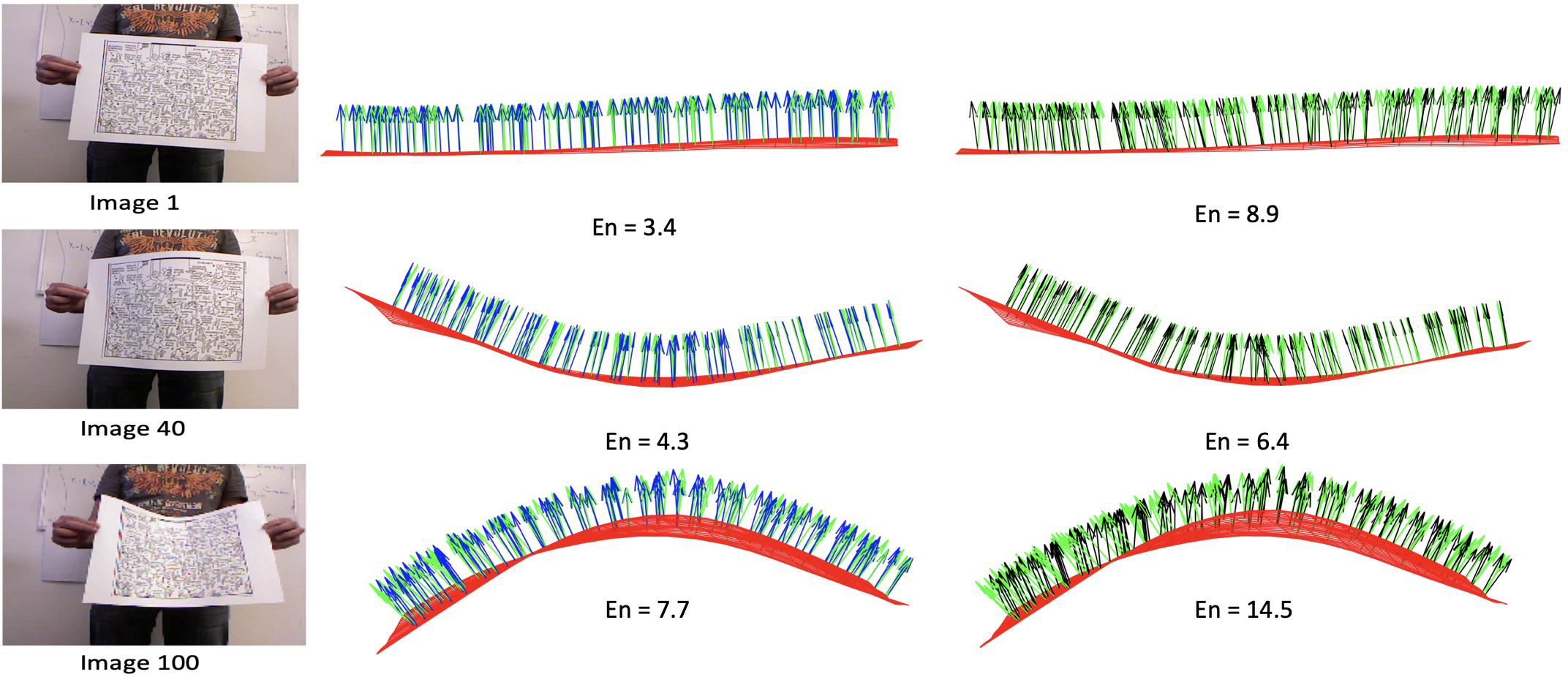}
      \includegraphics[width=\textwidth]{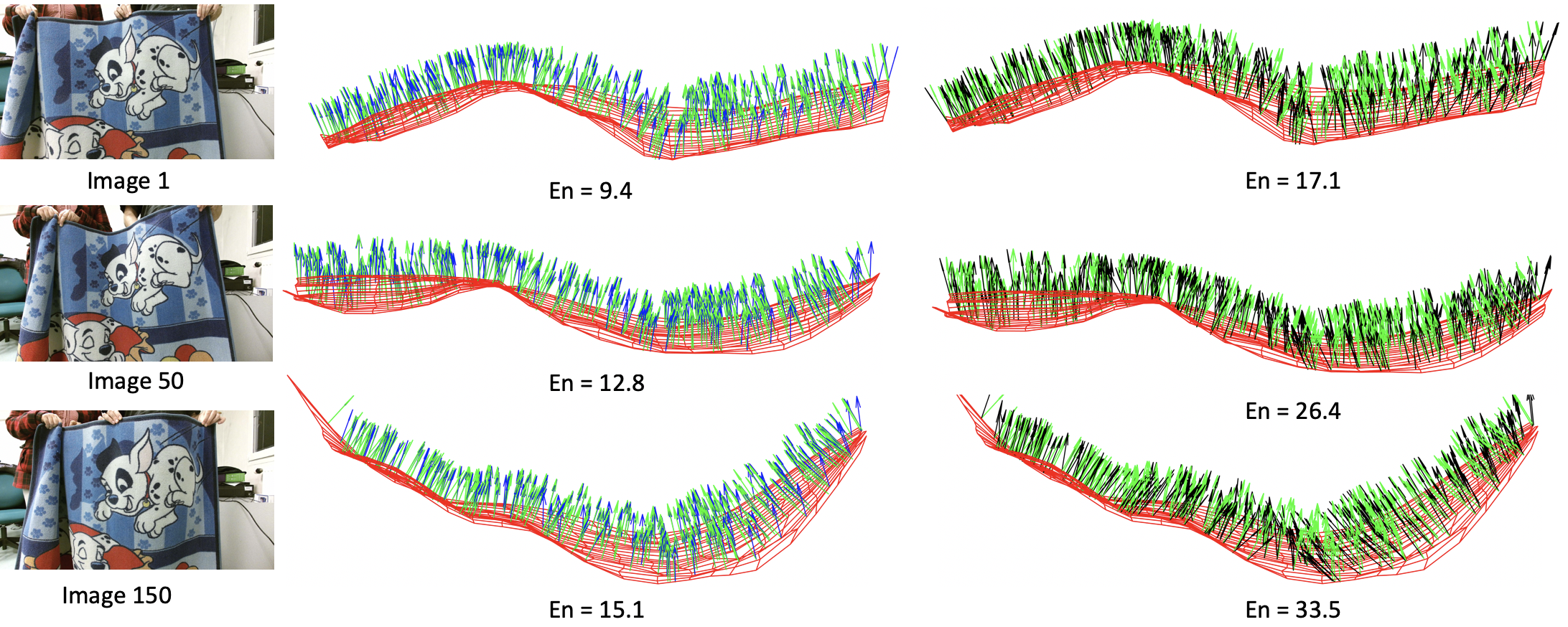}
      \includegraphics[width=\textwidth]{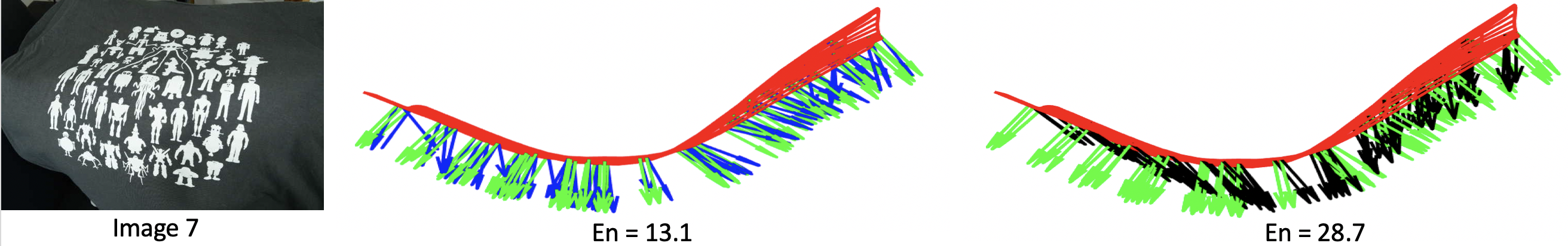}
     \caption{ {\bf Datasets used in our previous work.} Reconstructed normals on three images. The ground-truth normals are shown in green, the ones predicted by \textbf{Ours} in blue, and those by \textbf{Pa21-R} in black. Note that our normals are far less noisy. }
    \label{fig:paper_exp}
\end{figure*}



 \begin{table}
      \caption{{\bf Synthetic experiments results}. 'X' indicates that the method does not return a result because we are not using enough images. }
      \centering
     \includegraphics[width=0.5\textwidth]{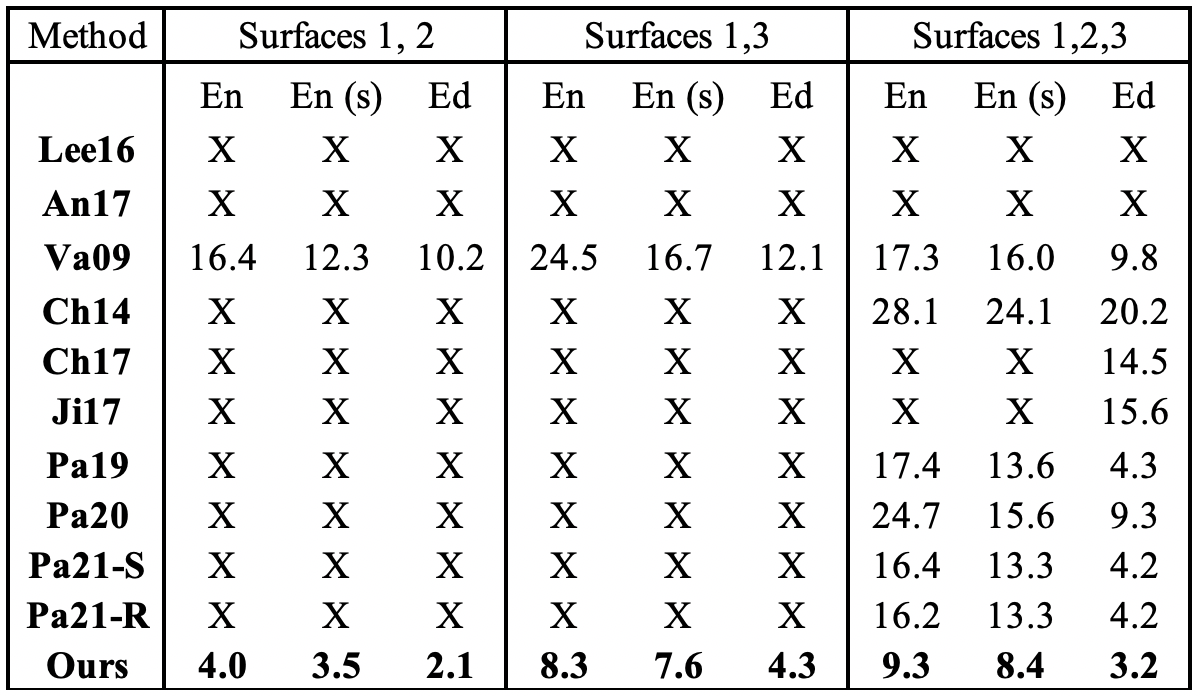}
        \label{fig:summary_syn}
    \end{table}

\subsection{Baselines and Metrics}

We compare  our method to local linearity-based diffeomorphic NRSfM \textbf{Pa20}~\cite{Parashar20}, jointly solving isometric/conformal NRSfM  \textbf{Pa19}~\cite{Parashar19a}, two fast solutions \textbf{Pa21-R} and \textbf{Pa21-S}~\cite{Parashar21} that transform the original constraints to univariate polynomials, which can be easily solved, and local and  piecewise homography decomposition,  \textbf{Ch14}~\cite{Chhatkuli14} and \textbf{Va09}~\cite{Varol09}, respectively. These are methods that, like ours, reconstruct local/piecewise surface normals and integrate them to obtain depth. Note that the solution to isometric NRSfM in~\cite{Parashar17} is the same as the one in \textbf{Pa19}. Therefore, there is no need for additional comparison. 

We report errors in terms of accuracy of the normals \textit{En} and 3D points \textit{Ed}. \textit{En} is computed as the average dot product between ground-truth and computed normals.  The normal integration done in the above methods yields a smooth reconstruction by enforcing a local smoothness on the normals. As a consequence, it improves the quality of the reconstructed normals. Therefore, we also report \textit{En (s)}, which is the error between the smoothened and the ground-truth normals.
\textit{Ed} is the mean RMSE between the ground-truth and computed 3D points. 

We also compare our approach against three of the best global methods, \textbf{Ch17}~\cite{Chhatkuli17}, \textbf{Ji17}~\cite{Ji17} and \textbf{Lee16}~\cite{Lee16}, along with a dense method, \textbf{An17}~\cite{Ansari17}. They directly return 3D points. Hence, we only report \textit{Ed} for these methods. 

While comparing on the datasets used by~\cite{Sidhu20}, we report $Ed$ as the mean 3D error, as computed in this method. Therefore, $Ed= \dfrac{1}{N}\sum_t \dfrac{|| P_{recon}-P_{GT} ||_2}{|| P_{GT}||_2}$, where $P_{recon}$ is the obtained reconstruction, $P_{GT}$ is the ground truth and $N$ is the number of images in the dataset. 

In the remainder of this section we will refer to the method described in this paper as~\textbf{Ours}.





  \begin{table*}
      \caption{ (left) {\bf RMSE results on the datasets used in our previous work}. 'X' indicates that the method does not evaluate normals. '----' indicates that method failed to return a result due to its high computational complexity. (right) {\bf Computation times} as a function of the number of images and points used.}
      \centering
     \includegraphics[width=\textwidth]{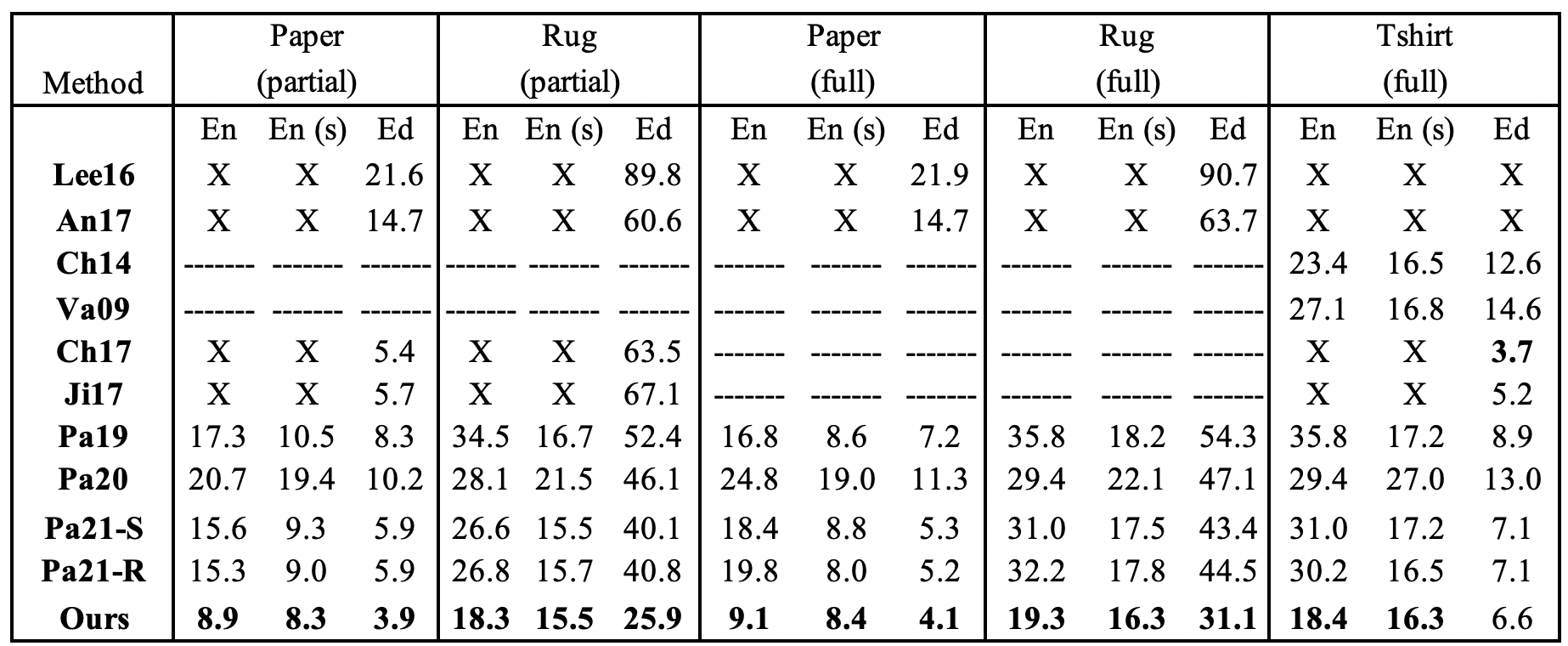}
        \label{fig:summary}
    \end{table*}

\subsection{Comparative Results}

\parag{Results on Synthetic Data.} Fig.~\ref{fig:syn_exp} shows  the generated surfaces. The performance of all  methods is averaged over 10 trials with added gaussian noise with a 3 pixels standard deviation.
 As \textbf{Ours} can reconstruct from two images only, we perform both pairwise reconstructions and joint reconstruction from the image triplet available for each surface. We report the results in Table~\ref{fig:summary_syn}. For methods that perform normal integration, we report errors of both computed and smoothened normals.  The improvement in the normals due to smoothing is huge for \textbf{Ch14} and \textbf{Va09}, substantial for \textbf{Pa19},  \textbf{Pa20}, \textbf{Pa21-S}  and \textbf{Pa21-R} and minor for our method. To truly compare the NRSfM techniques themselves,  we therefore report the accuracy of the computed normals rather than the smoothed ones. We obtain a very accurate reconstruction from 2 images only. Beside \textbf{Ours}, \textbf{Va09} is the only baseline that can reconstruct from 2 images. However, it does not perform well on this data.  \textbf{Lee16} and \textbf{An17} are designed for video sequences, and thus need more than 3 images to perform effectively.  The remaining methods can operate on three images, but their accuracy is lower than ours, especially in terms of normal accuracy. Since we can discard the normals that have a low reliability, the accuracy of our reconstruction is strengthened using multiple images. Fig.~\ref{fig:syn_exp} further confirms the quality of our reconstructions by depicting the normals we obtain {\it without} any smoothing.

 \begin{table}
      \caption{ {\bf Computation times} as a function of the number of images and points used.}
      \centering
     \includegraphics[width=0.5\textwidth]{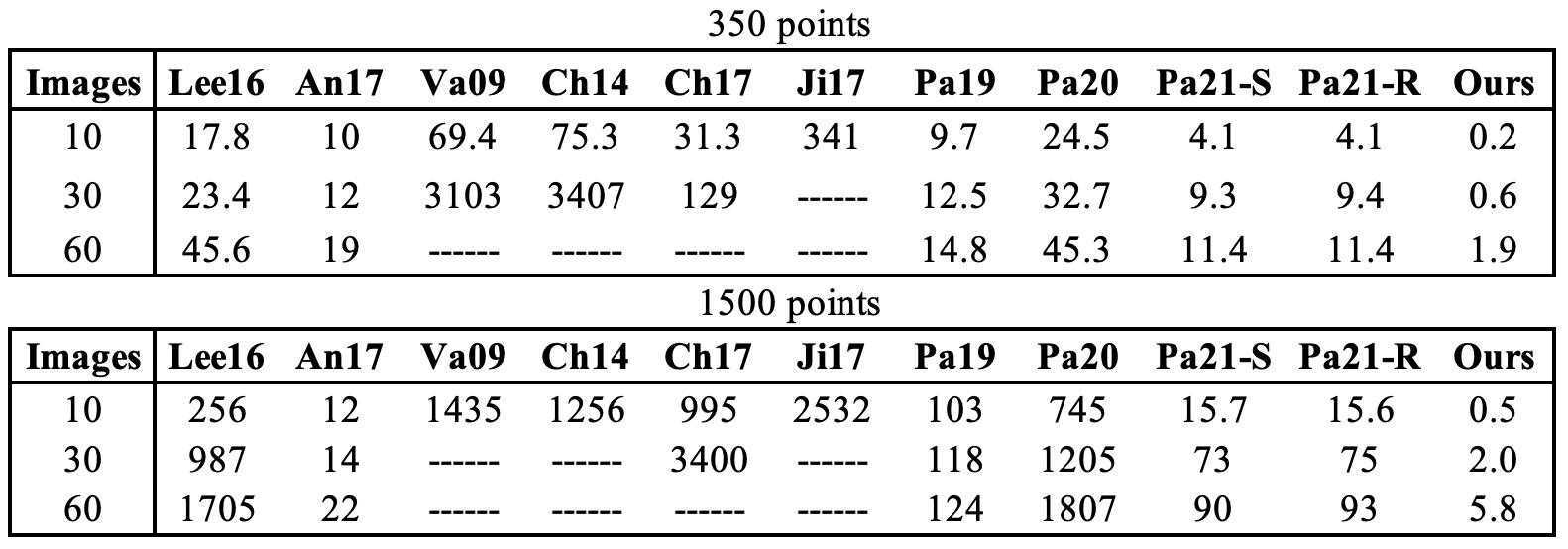}
        \label{fig:time}
    \end{table}

\parag{Results on the Datasets used in our Previous Work.}
Because the computational complexity of the global baselines grows rapidly with the number of correspondences, we evaluated all methods on the full set of correspondences and on a subset of 350 correspondences on {\bf Paper} and {\bf Rug}. For example, \textbf{Ch17}, \textbf{Ji17} have a cubic complexity and hence they yield a very high computation time when there are many correspondences. Their Matlab implementation crashes when using all correspondences, and using only 1000 correspondences still takes hours on a modern CPU.
Similarly \textbf{Ch14} and \textbf{Va09} take almost 1 hour to reconstruct 20 images and we therefore did not evaluate them on these datasets. The {\bf Tshirt} dataset has only 10 wide-baseline images.  \textbf{Lee16} and \textbf{An17} are not designed to work on wide-baseline data, therefore we did not evaluate them on this dataset. 

We report our quantitative results in Table~\ref{fig:summary}, and Figure~\ref{fig:paper_exp}  depict qualitative ones.  We outperform all baselines in terms of \textit{Ed} on the  \textbf{Paper} and \textbf{Rug} dataset with partial and full correspondences.  On the {\bf Tshirt} dataset, \textbf{Ch17} and \textbf{Ji17} perform better. Crucially, our performance is achieved at a much reduced computational cost by solving a set of equations in closed form, as opposed to invoking a complex solver. As a result, our approach is about 150 times faster than \textbf{Ch17} on 350 correspondences  and can handle thousands whereas \textbf{Ch17} cannot. Furthermore, our approach is also  50 times faster than \textbf{Pa19}, the counterpart local approach which uses expensive polynomial solvers, because we do not have to derive a complicated formulation to obtain a unique solution for each correspondence. 

Table~\ref{fig:time}  provides a detailed analysis of the run-times of all the methods on 350 and 1500 points. We assume that the input point correspondences and their derivatives are pre-computed. Therefore, the timings only encode the computation of the normals or 3D points. Our approach yields the fastest run-times, seconded by \textbf{An17}. Note, however, that \textbf{An17} has a parallel implementation and is computationally optimized. By contrast, our approach, as all the other ones, is implemented in Matlab and not optimized for speed.

The relative slowness of the other local method arises from the local normal estimators of \textbf{Pa19} and \textbf{Pa20} having to minimize the sum of squares of polynomials,
which is expensive even if it has linear complexity. \textbf{Pa20} is further slowed down by having to transform polynomials into univariate expressions. \textbf{Pa21-S} and \textbf{Pa21-R} obtain analytical solutions but require a fairly expensive disambiguation. By contrast, our local normal estimator is computationally cheap as it has a closed-form solution. 

\begin{figure*}
    \centering
     \includegraphics[width=\textwidth]{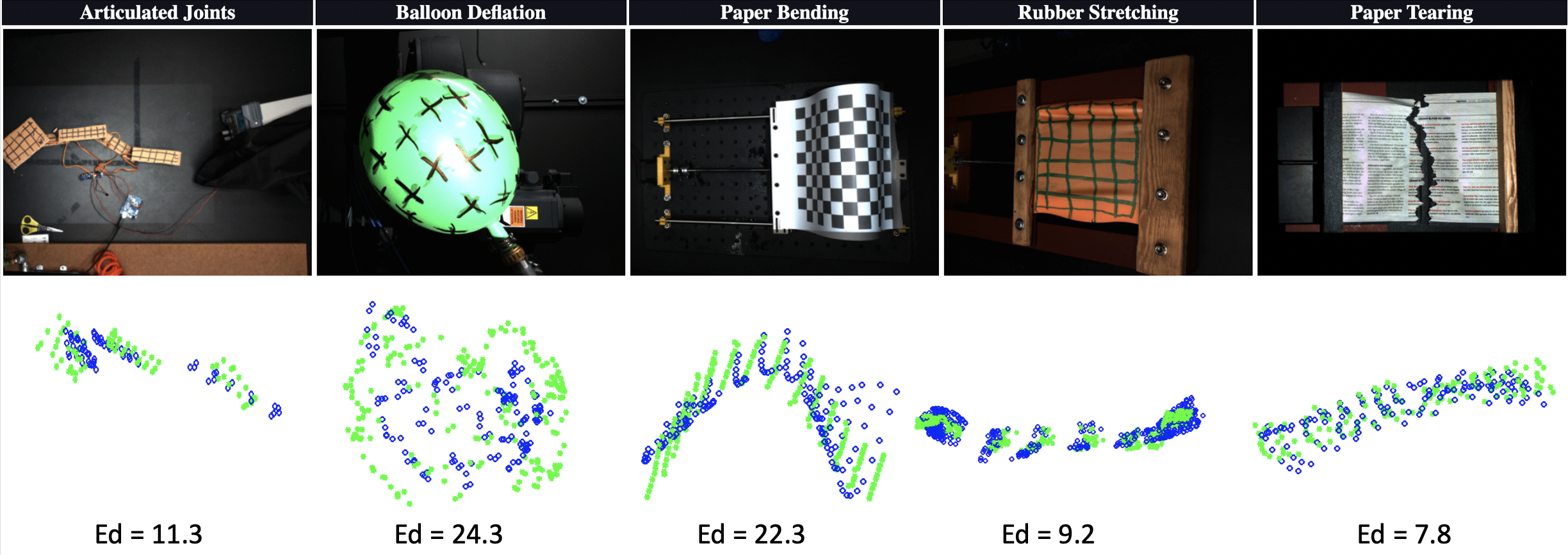}
    \caption{NRSfM challenge dataset and some reconstructions using {\bf Ours}. Green indicates the ground truth and blue indicates our reconstruction.}\label{fig:nrsfm_ch_data}
\end{figure*}

\parag{Results on the NRSfM Challenge Dataset.}
 Fig.~\ref{fig:challenge} compares the performance of \textbf{Ours}  with that of other methods in terms of $Ed$, measured in mm, with {\bf Best} being the one that does best as reported in the benchmark statistics provided on the website. The local methods show a significant performance improvement compared to the other ones. \textbf{Pa19} uses second-order derivatives of the image registration $\eta$, which can be highly erroneous on this dataset.  It uses an expensive polynomial solver, which cannot handle such large noise and fails on a large number of cases.  \textbf{Pa21-S} and \textbf{Pa21-R} find an analytical solution to the isometric/conformal NRSfM posed in \textbf{Pa19}, which requires a non-linear refinement to obtain a unique solution; they obtain decent results on this dataset. \textbf{Pa20} solves NRSfM using diffeomorphic constraints, which uses only first-order derivatives of $\eta$, and is thus less impacted by the sparsity of the data and performs better than \textbf{Pa21-S} and \textbf{Pa21-R}. \textbf{Ours} requires second-order derivatives of the image registration, but it is equipped with a measure to compute the well-conditioning of the data. This lets us identify and discard the non-isometric/non-conformal data and reconstruct from  as-isometric(or conformal)-as-possible data. As a result, \textbf{Ours} yields better results than \textbf{Pa20}. Fig.~\ref{fig:nrsfm_ch_data} shows some reconstructions obtained with our method. 

 \begin{table*}
      \caption{ {\bf Results on the NRSfM challenge datasets.}}
      \centering
     \includegraphics[width=\textwidth]{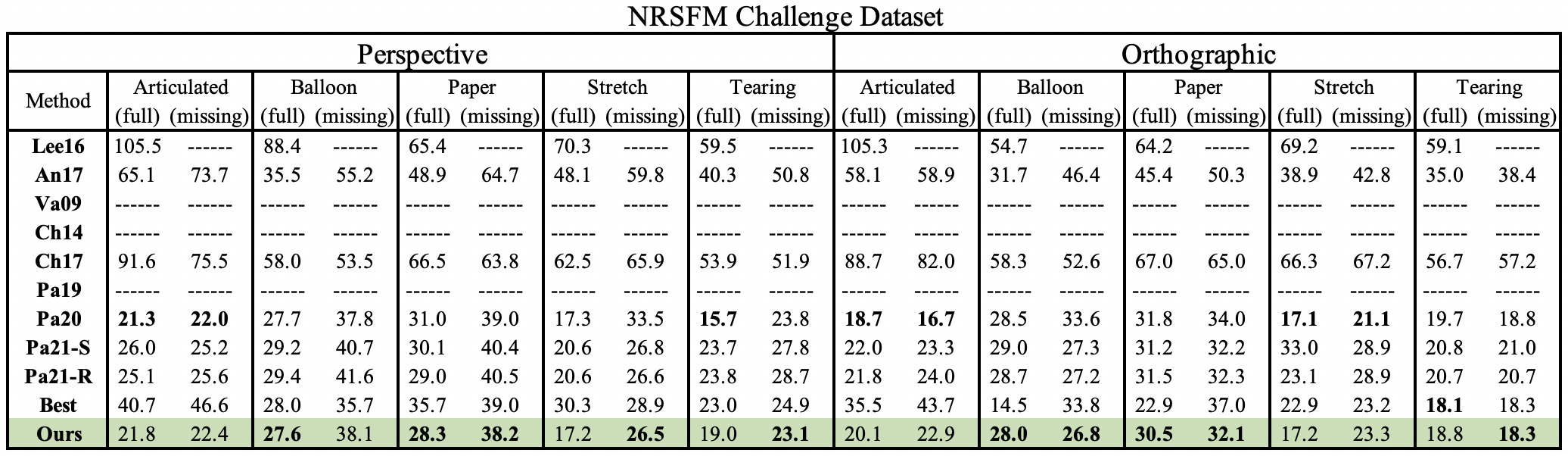}
        \label{fig:challenge}
    \end{table*}

\parag{Results on the Blue Sheet Dataset and on the Datasets used by~\cite{Sidhu20}.} 
These datasets are large in terms of the number of either point correspondences or images they contain. We compare the performance of \textbf{Ours} with \textbf{An17}, which is designed for reconstructing dense objects, however, it takes several hours to reconstruct. Additionally, we report the performance of our other local methods \textbf{Pa19}, \textbf{Pa21-S} and \textbf{Pa21-R}. In this case, we report the mean 3D error to be able to compare with the performance of~\cite{Sidhu20}, which has demonstrated best results on these datasets.  Table~\ref{fig:dense} summarizes the results.  \textbf{Ours} performs better than most of the methods on these datasets. The \textbf{Actor} and \textbf{Expressions} sequences are relatively simple, with small relative motion across images. All local methods therefore perform similarly on these sequences.  \textbf{An17} performs better than \textbf{Ours} on the \textbf{Actor} sequence. However, the visual performance  is quite similar as the error margin is very low, to the third decimal place. Fig.~\ref{fig:ac_exp} shows some reconstructions.  Fig.s~\ref{fig:sheet_exp}~\ref{kin_exp} show the results on the \textbf{ Blue Sheet}, \textbf{Paper} and \textbf{Tshirt} datasets where \textbf{Ours}  performs significantly better than the compared methods.

 \begin{figure*}
    \centering
     \includegraphics[width=\textwidth]{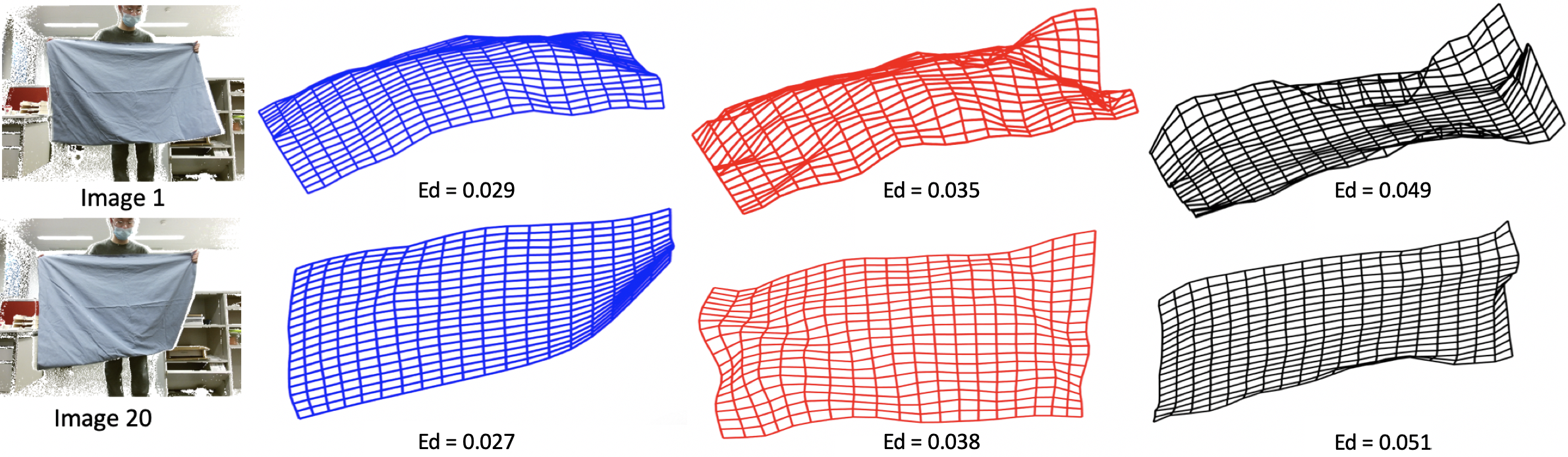}
     \caption{ {\textbf{Blue Sheet dataset.}} Reconstructed surfaces for two images. The predictions of \textbf{Ours} are shown in blue, of \textbf{Pa21-R} in red, and of \textbf{An17}  in black. Note that our reconstructions are less noisy and match the surface 3D shape much better. }
    \label{fig:sheet_exp}
\end{figure*}


     \begin{table}
      \caption{ {\bf Performance on dense datasets.} }
      \centering
     \includegraphics[width=0.5\textwidth]{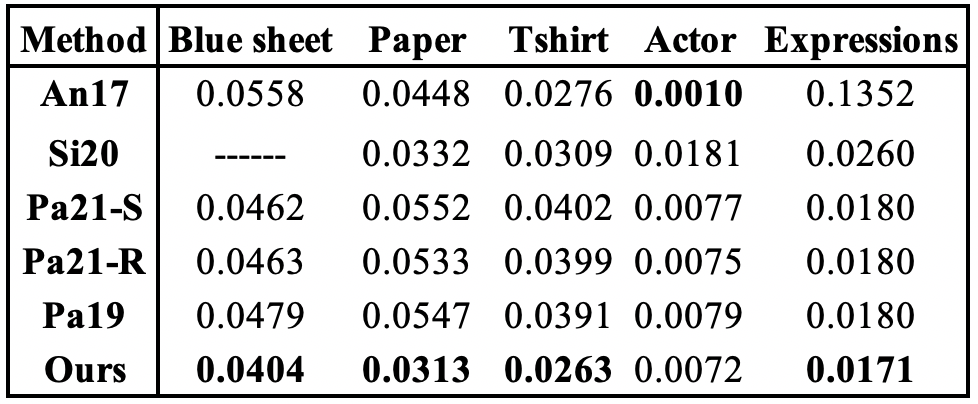}
        \label{fig:dense}
    \end{table}

     \begin{figure}
    \centering
     \includegraphics[width=0.5\textwidth]{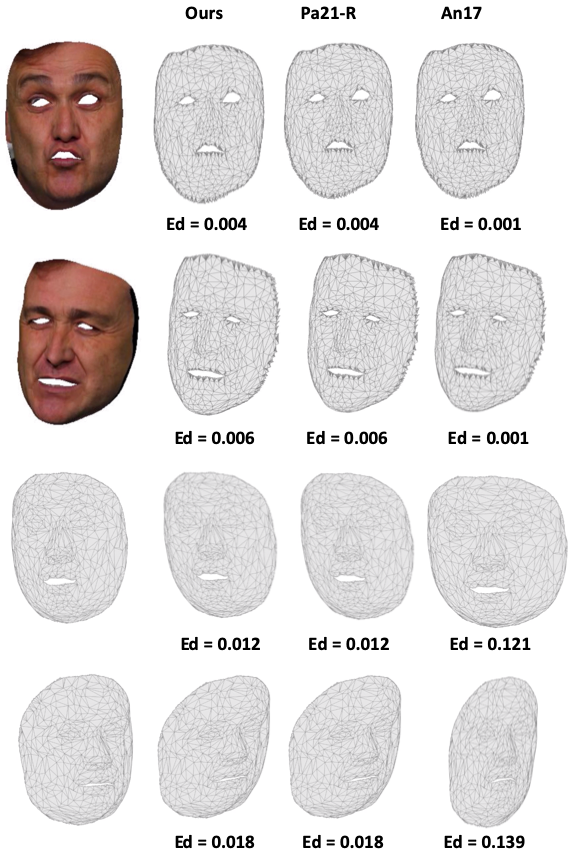}
     \caption{ {\textbf{Actor and Expressions datasets.}} Reconstructed surfaces for two images. The predictions of \textbf{Ours}, \textbf{Pa21-R} and of \textbf{An17} are quite similar.}
    \label{fig:ac_exp}
\end{figure}

 \begin{figure}
    \centering
     \includegraphics[width=0.5\textwidth]{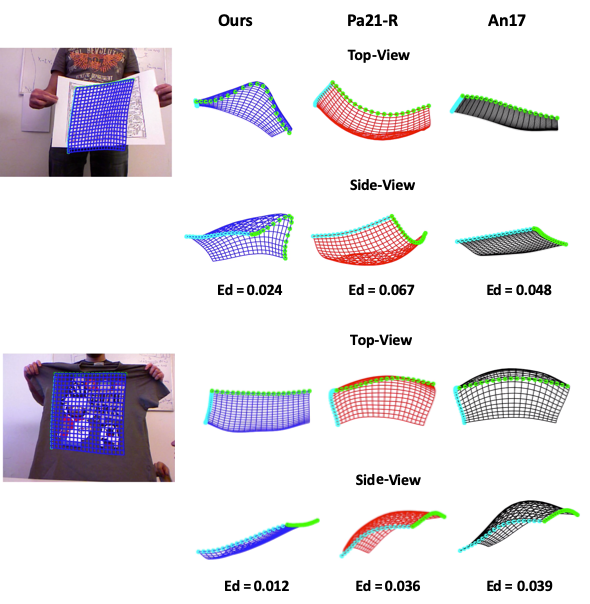}
     \caption{ {\textbf{Paper and Tshirt datasets.}} The predictions of \textbf{Ours} are shown in blue, of \textbf{Pa21-R} in red, and of \textbf{An17}  in black. Note that our reconstructions are less noisy and match the surface 3D shape much better. }
    \label{fig:kin_exp}
\end{figure}


\section{Conclusion}

We have proposed an approach to NRSfM that can estimate normals from image pairs given a 2D warp and point correspondences between the two images. It does so in closed form from individual correspondences and is therefore fast. Furthermore, it can estimate if these normals are reliable given the motion from one image to the next. When they are found to be, our experiments show that they are indeed very accurate. As a result, our method performs well with various deformation types and can reconstruct large and small  deformations at a low  computational cost.
Our next step will be to remove the dependency on expensive methods to compute warps and integrate normals so that a truly real-time application can be developed.

\bibliographystyle{ieee}
\bibliography{string,vision,geom}

\end{document}